\documentclass[conference]{IEEEtran}
\IEEEoverridecommandlockouts

\usepackage{cite}
\usepackage{amsmath,amssymb,amsfonts}
\usepackage{newtxtext,newtxmath}
\usepackage{algorithmic}
\usepackage{graphicx}
\usepackage{textcomp}
\usepackage{xcolor}
\usepackage{booktabs}
\usepackage{float}
\usepackage{cleveref}
\def\BibTeX{{\rm B\kern-.05em{\sc i\kern-.025em b}\kern-.08em
    T\kern-.1667em\lower.7ex\hbox{E}\kern-.125emX}}
\begin{document}

\title{Efficient 4D Gaussian Stream with Low Rank Adaptation*\\
}

\author{
    \IEEEauthorblockN{Zhenhuan Liu, Shuai Liu, Yidong Lu}
    \IEEEauthorblockA{\textit{Dept. of Automation} \\
    \textit{Shanghai Jiao Tong University} \\
    Shanghai, China \\
    jimliu20200918@gmail.com, shuailiu@sjtu.edu.cn, yidonglu116@gmail.com}
    \hfill
    \IEEEauthorblockN{Yirui Chen, Jie Yang, Wei Liu}
    \IEEEauthorblockA{\textit{Dept. of Automation} \\
    \textit{Shanghai Jiao Tong University} \\
    Shanghai, China \\
    \{chenyirui, jieyang, weiliucv\}@sjtu.edu.cn}
}
\maketitle

\begin{abstract}
Recent methods have made significant progress in synthesizing novel views with long video sequences. This paper proposes a highly scalable method for dynamic novel view synthesis with continual learning. 
We leverage the 3D Gaussians to represent the scene and a low-rank adaptation-based deformation model to capture the dynamic scene changes. Our method continuously reconstructs the dynamics with chunks of video frames, reduces the streaming bandwidth by $90\%$ while maintaining high rendering quality comparable to the off-line SOTA methods. 
\end{abstract}


\section{Introduction}

Novel view synthesis (NVS) is a fundamental problem in computer vision. It aims to generate novel views of a scene from a set of input images. The problem has been widely studied in the literature and has many applications, such as autonomous driving. Recent years have seen a surge of interest in dynamic novel view synthesis, where the input images are video frames instead of static images. Current solutions for the dynamic NVS problem could be divided into two categories: the offline methods and the online methods. The traditional offline methods load all the input images into memory (or store them onto disk) and generate video frames in a batch manner. 
The offline methods could achieve high-quality results but requir a large amount of memory or disk storage for caching the video frames. The demand for memory or disk storage is proportional to the number of input images, which makes the offline methods infeasible for long video sequences. 
The recent online methods, on the other hand, are usually more memory-efficient but suffer from the quality degradation due to the limited access to the input images. 

Although current methods like 3DGStream \cite{3DGStream} and CD-NGP \cite{cdngp} have made significant progress in dynamic novel view synthesis, they still have limitations in terms of memory efficiency and rendering quality, respectively. To address these limitations, we propose a 3D Gaussian-based method with low-rank adaptation for dynamic novel view synthesis. 

\section{Related Work}

\subsection{Dynamic Novel View Synthesis}

The dynamic novel view synthesis problem has been widely studied in the literature. Current methods could be divided into two categories: the neural radiance field (NeRF)-based methods and the 3D Gaussian-based methods. The NeRF-based methods represent the scene as a continuous 5D function and achieve high-quality results. However, the pixel-based volume rendering are computationally expensive and are hard in achieving real-time rendering. 
The 3D Gaussian-based methods, on the other hand, represent the scene as a set of 3D Gaussians and are inherently suitable for real-time rendering. 

D-NeRF \cite{Dnerf} employs a large deformation MLP to capture the intricate coordinate transitions in dynamic simulated environments. Similarly, DyNeRF \cite{Dynerf} encodes temporal data using long vectors and decodes the features into radiance fields with large MLPs, enabling high-quality novel view synthesis in real dynamic scenes. In \cite{park2023temporal}, an interpolation-based MLP representation for dynamic scenes is introduced, offering a significant speedup over DyNeRF. 

In the realm of tensor-based representations,  K-Planes \cite{kplanes} and HexPlane \cite{HexPlane_} decompose the 4D field $(x,y,z,t)$ into the product of 2D planes, achieving similar high-fidelity results with an intermediate model size (200MB) and training time (100 minutes) on the DyNeRF dataset.  HyperReel \cite{hyperreel} uses a network to predict sample locations and models dynamic scenes via a key-frame-based TensoRF \cite{tensorf} representation, ensuring high fidelity and real-time rendering without the need for custom CUDA kernels.

The third category encompasses  voxel-based representations. TineuVox \cite{tineuvox} utilizes the deformation MLP from D-NeRF and incorporates a 3D voxel grid with multi-scale interpolation, resulting in high fidelity and fast convergence on the D-NeRF dataset. On the other hand, MixVoxels \cite{mixvoxels} calculates a variation field by measuring the standard deviation of all pixels over time, allowing it to decompose dynamic scenes into static and dynamic voxels. This approach, with its distinct representations for static and dynamic objects, achieves both high-fidelity results and extremely fast convergence (15 minutes) on the DyNeRF dataset. Additionally, \cite{nerfplayer} trains Instant-NGP \cite{ingp} and TensoRF \cite{tensorf} offline, streaming the models through feature channels for dynamic scene rendering. CD-NGP \cite{cdngp} splits the scene into multiple chunks and train different branches accordingly. They use a large base hash table to store the spatial features of the scene and use small auxiliary hash tables to cope with feature transitions, achieving high-quality results with a small memory footprint.

4D Gaussian point clouds \cite{4Dgaussians,gs4dhust} and 3D Gaussians with deformation networks \cite{deformable3Dgaussians} are employed for real-time, high-fidelity rendering of dynamic scenes.

\section{Method}
We propose our method (LR-4DGStream) based on the previous SOTA method E-D3DGS \cite{ed3dgs}. Our method comprises three main aspects: using the 3D Gaussian Splatting for real-time rendering, using the plane-based deformation module for capturing the dynamic scene changes, and using the low-rank adaptation for enhancing scalability and reducing the streaming bandwidth. 

\subsection{3D Gaussian point clouds}
We use the 3D Gaussian point clouds to represent the scene details. 
Each 3D Gaussian contain the following properties: Gaussians center $\boldsymbol{\mu} \in \mathbb{R}^3$, opacity $o\in \mathbb{R}^{+}$, color $\boldsymbol{c}$ and a covariance matrix $\mathbf{\Sigma} \in \mathbb{S}^{++}$. In implementation,  $\Sigma$ is decomposed into a rotation matrix $\mathbf{R}$ and a scaling matrix $\mathbf{S}$, i.e. $\Sigma=\mathbf{RSS}^{\top}\mathbf{R}^{\top}$. The view-dependent colors $c$ are represented by spherical harmonics. 

Following~\cite{zwicker2001ewa}, the 3D Gaussian primitives can be projected from world coordinates to the image plane by the 2D covariance matrix $\mathbf{\Sigma}^{\prime}$:
\begin{equation}
\mathbf{\Sigma}^{\prime}=\mathbf{J}\mathbf{W}\mathbf{\Sigma}\mathbf{W}^{\top}\mathbf{J}^{\top},
\end{equation}
where $\mathbf{J}$ is the Jacobian of the affine transformation and $\mathbf{W}$ is the world-to-camera transform matrix.
3DGS is suitable for differentiable rendering with the following blending function:
\begin{equation}
    \mathbf{C}(\mathbf{p})=\sum_{i\in \mathbb{N}}\mathbf{T}_i \mathbf{\alpha}_ic_i,\quad \mathbf{\alpha}_i=o_i e^{-\frac12(\mathbf{p}-\mu_i)^T \mathbf{\Sigma} ^{\prime}(\mathbf{p}-\mu_i)},
    \label{eq:blending}
\end{equation}
where $\mathbf{\alpha}_i$ denotes the alpha values, $c_i$ is the view-dependent appearance and $\mathbf{T}_{i}= {\textstyle \prod_{j=1}^{i-1}} (1-\alpha_{j})$ is the transmittance.

\subsection{Deformation module}
We apply the per-gaussian embedding to represent detailed dynamics in the scene. In addition, we use a tri-plane representation to encode the local features of the scene. Let the parameters of scaling matrix and rotation matrix be $S\in \mathbb{R}^{3}, R\in \mathbb{R}^4$, the spherical harmonics (SH) coefficients of the Gaussians be $S_c$, The per-gaussian embeddings are $\eta \in \mathbb{R}^{d_e}$.

\subsection{Low Rank adaptation and continual learning}

In order to reduce the streaming bandwidth, 
we leverage the low-rank adaptation to adapt the plane encoder of the deformation module. The low-rank \emph{adaptation} is a technique to fine-tuning a matrix by another matrix with lower rank. Consider a matrix $A\in \mathbb{R}^{m\times n}$ with SVD decomposition $A = U\Sigma V^T$, where $U\in \mathbb{R}^{m\times m}$, $\Sigma\in \mathbb{R}^{m\times n}$, and $V\in \mathbb{R}^{n\times n}$ are the left singular vectors, singular values, and right singular vectors of $A$, respectively. 
The $\lambda$-rank adaptation of $A$ is given by :
\begin{equation}
    A_\lambda = U_\lambda\Sigma_\lambda V_\lambda^T,
\end{equation}
\noindent where $U_k\in \mathbb{R}^{m\times \lambda}$, $\Sigma_k\in \mathbb{R}^{\lambda \times \lambda}$, and $V_k\in \mathbb{R}^{n\times \lambda}$ are the left singular vectors, top-$\lambda$ singular values, and right singular vectors of $A$, respectively. Compared with another matrix $B\in \mathbb{R^{M\times N}}$ with full rank,  the low-rank adaptation could reduce the storage and computation cost of $A$ by $O(mn - \lambda(m+n))$. 
For the simplicity of implementation, we set $\Sigma_k$ as the identity matrix. Thus, the $\lambda$-rank adaptation of $A$ is given by: 
\begin{equation}
    A_\lambda = U_\lambda V_\lambda^T. 
\end{equation}
\noindent where $U_k \in \mathbb{R^{M\times \lambda}}$, $V_k \in \mathbb{R^{N\times \lambda}}$ are the left and right adaptation vectors of $A$, respectively. 

Compared with the previous hash-encoding-based method \cite{cdngp}, the low-rank adaptation achieves low streaming bandwidth and continual  adaptability simultaneously.

Although the previous hash encoding-based method significantly reduces the bandwidth by reducing the numbers of the elements of the hash table, it falls short in continually adapting to new time intervals due to the coordinate-based hashing process.  Let $\Psi_0(x)$ be the large hash table of the base branch, $\Psi_k(x)$  be the small hash tables of the auxiliary branches. The hash encoding of the $k$-th branch is $\Psi^{(k)}(x) = \Psi_0(x) + \Psi_k(x)$. On the one hand, the simple feature fusion process omits the intermediate hash tables, which makes it unable to handle large feature transitions between the base branch and the $k$-th branch.  On the other hand, the summation of every hash table in the intermediate branches and the auxiliary branches, i.e. $\Psi^{(k)}(x) = \Psi_0(x) + \sum_{j=1}^{k-1} \Psi_j(x)$ is computationally expensive. The disadvantage derives from the grid-coordinate-based hashing process and the fused CUDA kernels, which makes it difficult to establish a mapping between the  elements of the base hash table and the auxiliary hash tables.

On the contrary, the low-rank adaptation could feasibly adapt to new time intervals by simply performing the summation of the original matrix and the adaptation matrix, i.e. $A^{(k)} = A^{(k-1)} + U_{\lambda}^{(k-1)} V_{\lambda}^{(k-1)}$. This property enables the low-rank adaptation to continually adapt to the difference between the $k$-th and the $(k-1)$-th branch.

\newcommand{\uaa}{$\uparrow$}
\newcommand{\daa}{$\downarrow$}

\begin{table*}[!ht]
    \centering
    \caption{Comparisons between our method and state-of-the-art  methods on the DyNeRF dataset. Memory denotes the maximum host memory required during training. $\dagger$ denotes the methods evaluated only on the \emph{flame salmon} scene. 
     \emph{-} denotes the metrics not reported in the original publications. }
     \begin{tabular}{l|ccllllllr}  
      \toprule
    Method    & Online & Offline   & $T_{chunk}$     & PSNR\uaa   & DSSIM\daa & LPIPS\daa  & Time\daa             & Size\daa  & Memory\daa \\ 
  \midrule
  DyNeRF$^{\dag}$ \cite{Dynerf} & & $\checkmark$ & 300                       & 29.58  &  0.020 &   0.083  & 7 days  &  28MB     &  >100GB      \\
  NeRF-T$^{\dag}$ \cite{Dynerf}  & & $\checkmark$  & 300                   & 28.45  &   0.023 & 0.10   & -                & -        &  >100GB       \\
  Interp-NN \cite{park2023temporal}  & & $\checkmark$  & 300            & 29.88  & -     & 0.096 & 2 days      & \textbf{20MB}     &  >100GB       \\
  HexPlane \cite{HexPlane_}     & & $\checkmark$  & 300               &   31.57  & \textbf{0.016} & 0.089 &   96 min           & 200MB    &  62GB       \\
  K-Planes \cite{kplanes}       & & $\checkmark$  & 300             &  31.63  & -     & -      & 110 min          & 200MB    &  >100GB       \\
  MixVoxels \cite{mixvoxels}  & & $\checkmark$  & 300              & 30.71  & 0.024 & 0.162 & \textbf{15 min}            & 500MB    &  >100GB      \\        
  4DGS \cite{4Dgaussians}  & & $\checkmark$  & 300              & \textbf{32.01}   & - & \textbf{0.055} & 6 hours            & 4000MB    &   \textbf{<10GB}    \\   
  HyperReel \cite{hyperreel}  & & $\checkmark$  & 50          & 31.1  & - & 0.096 & 9 hours            & 360MB    &      18GB    \\  
  NeRFplayer \cite{nerfplayer}  & & $\checkmark$  & 300               & 30.29  & - & 0.152  & 5.5 hours            & -     &  >100GB   \\ \midrule 
  StreamRF \cite{streamRF} & $\checkmark$ & & 1              & 28.85  & 0.042     & 0.253      &  75 min           & 5000MB   & \textbf{<10GB}       \\
  HexPlane-PR & $\checkmark$ & & 10  & 24.03 & 0.081 &  0.244  & 100 min            & 318MB    &    14GB     \\
  HexPlane-GR & $\checkmark$ & & 10  & 25.17 & 0.050  & 0.272    & 125 min            & 318MB    &    14GB     \\                          
  INV \cite{invnerf} &  $\checkmark$ & & 1             & 29.64  &   0.023 &  0.078 & 40 hours            & 336MB     &  \textbf{<10GB}     \\
  CD-NGP & $\checkmark$ & & 10        & 30.23  & 0.027& 0.198 &  75 min            &    113MB     &    14GB      \\
  LR-4DGStream (Ours) & $\checkmark$ & & 50 & 30.27 & 0.017 & 0.058 &  120 min & 160MB & 50GB \\
  
   \bottomrule  
  \end{tabular}
     \label{tab:comparison_dynerf_dataset}
  \end{table*}

\section{Experiments and analysis}
\subsection{Dataset and evaluation metrics}
The DyNeRF dataset \cite{Dynerf} serves as the primary benchmark for dynamic view synthesis, offering high-quality multi-camera videos that are widely used to evaluate leading-edge methods. This dataset comprises six distinct cooking scenes, each showcasing diverse lighting conditions, changes in food topology, and transient effects like flames. It includes five videos with 300 frames each and one extended video with 1200 frames, all recorded at a resolution of $2704\times2028$ with up to 21 camera angles at 30 frames per second (FPS). Following the majority of the baselines, we use the initial 300 frames of the 1200-frame \emph{flame salmon} video for assessment. 

We evaluate the reconstruction quality of the methods using the most prevailing metrics in the field: PSNR, DSSIM, and LPIPS \cite{lpips}.

\subsection{Implementation details}
Our implementation is based on the previous SOTA method E-D3DGS. 
Following the majority of existing methods, we use 
$1352\times 1014$ resolution for training and evaluation. We use AlexNet \cite{alexnetcite} to compute LPIPS metric.

All metrics are obtained using one NVIDIA RTX 3090 GPU if not specified otherwise.

\subsection{Results and Comparison}
We compare our method with the state-of-the-art methods on the DyNeRF dataset. The results are shown in Table \cref{tab:comparison_dynerf_dataset}. Our method outperforms CD-NGP by a significant margin. Moreover, our method supports 20FPS rendering. When applied to streaming scenario, our method only consumes 13MB/chunk (0.26MB/frame), which is $33\%$ less than the recent hash-based CD-NGP method.

\subsection{Ablation study}

We conduct ablation studies and show the results in \cref{tab:ablation}. 

\begin{table}
  \caption{Ablation study 
  \label{tab:ablation}}
  \centering
\begin{tabular}{llccc}
  \toprule
  {Hidden dims} &      Freq. Enc.         & {PSNR $\uparrow$} & {DSSIM $\downarrow$} & {LPIPS $\downarrow$} \\ \midrule
  $256$        &               & \textbf{30.27}        & \textbf{0.017}        & \textbf{0.058}       \\ 
  $512$     &                & 29.81        & {0.025}        & 0.077       \\    
  \bottomrule
\end{tabular}
\end{table}

\section{Conclusion and limitation}

We propose LR-4DGStream, a highly scalable method for dynamic novel view synthesis under continual learning setting. Our method leverages the 3D Gaussians to represent the scene and a low-rank adaptation-based deformation model to capture the dynamic scene changes. We use per-gaussian embedding and low-rank adaptation-based plane representation to reduce the streaming bandwidth by $90\%$ while maintaining high rendering quality comparable to the off-line SOTA methods. However, like most existing methods for dynamic novel view synthesis, our method falls short in generalizing across different scenes, which is a challenging problem in the field. Future work will focus on improving the generalization ability of our method and extending it to more complex scenes.

\bibliographystyle{IEEEtran}
\bibliography{refs}

\vspace{12pt}

\end{document}